\documentclass[times, twoside, watermark]{zHenriquesLab-StyleBioRxiv}
\usepackage{blindtext}

\leadauthor{Schmidgall} 

\begin{document}

\title{Brain-inspired learning in artificial neural networks: a review}
\shorttitle{Brain-inspired learning in ANNs: a review}

% Use letters for affiliations, numbers to show equal authorship (if applicable) and to indicate the corresponding author
\author[1, \Letter]{Samuel Schmidgall}
\author[2,3]{Jascha Achterberg}
\author[4]{Thomas Miconi}
\author[5]{Louis Kirsch}
\author[6]{Rojin Ziaei}
\author[6]{S. Pardis Hajiseyedrazi}
\author[7]{Jason Eshraghian}
%\author[1]{Andreas Andreou} :(

\affil[1]{Johns Hopkins University}%, Department of Electrical and Computer Engineering}
\affil[2]{University of Cambridge}%, MRC Cognition and Brain 
\affil[3]{Intel Labs}%, MRC Cognition and Brain Sciences Unit}
\affil[4]{ML Collective}
\affil[5]{The Swiss AI Lab IDSIA}%, USI, SUPSI}
%\affil[4]{Columbia University, Center for Theoretical Neuroscience}
\affil[6]{University of Maryland, College Park}%, Department of Information Studies}
%\affil[6]{University of Maryland}%, College Park, Department of Electrical and Computer Engineering}
\affil[7]{University of California, Santa Cruz}%, Santa Cruz, Department of Electrical and Computer Engineering}

\maketitle

%TC:break Abstract
%the command above serves to have a word count for the abstract
\begin{abstract}
Artificial neural networks (ANNs) have emerged as an essential tool in machine learning, achieving remarkable success across diverse domains, including image and speech generation, game playing, and robotics. However, there exist fundamental differences between ANNs' operating mechanisms and those of the biological brain, particularly concerning learning processes. This paper presents a comprehensive review of current brain-inspired learning representations in artificial neural networks. We investigate the integration of more biologically plausible mechanisms, such as synaptic plasticity, to enhance these networks' capabilities. Moreover, we delve into the potential advantages and challenges accompanying this approach. Ultimately, we pinpoint promising avenues for future research in this rapidly advancing field, which could bring us closer to understanding the essence of intelligence.
\end {abstract}
%TC:break main
%the command above serves to have a word count for the abstract

%\begin{keywords}
%brain-inspired learning | artificial ne | bla | bla
%\end{keywords}

\begin{corrauthor}
%\texttt{r.henriques{@}ucl.ac.uk}
\texttt{sschmi46@jhu.edu}
\end{corrauthor}

\section*{Introduction}

The dynamic interrelationship between memory and learning is a fundamental hallmark of intelligent biological systems. It empowers organisms to not only assimilate new knowledge but also to continuously refine their existing abilities, enabling them to adeptly respond to changing environmental conditions. This adaptive characteristic is relevant on various time scales, encompassing both long-term learning and rapid short-term learning via short-term plasticity mechanisms, highlighting the complexity and adaptability of biological neural systems \cite{newell2001time, stokes2015activity, gerstner2018eligibility}. The development of artificial systems that draw high-level, hierarchical inspiration from the brain has been a long-standing scientific pursuit spanning several decades. While earlier attempts were met with limited success, the most recent generation of artificial intelligence (AI) algorithms have achieved significant breakthroughs in many challenging tasks. These tasks include, but are not limited to, the generation of images and text from human-provided prompts\cite{beltagy2019scibert,brown2020language, ramesh2022hierarchical, saharia2022photorealistic}, the control of complex robotic systems\cite{kumar2021rma, miki2022learning, fu2022deep}, and the mastery of strategy games such as Chess and Go\cite{silver2018general} and a multimodal amalgamation of these \cite{driess2023palm}.

\begin{figure*}
    \centering
\includegraphics[width=0.95\linewidth]{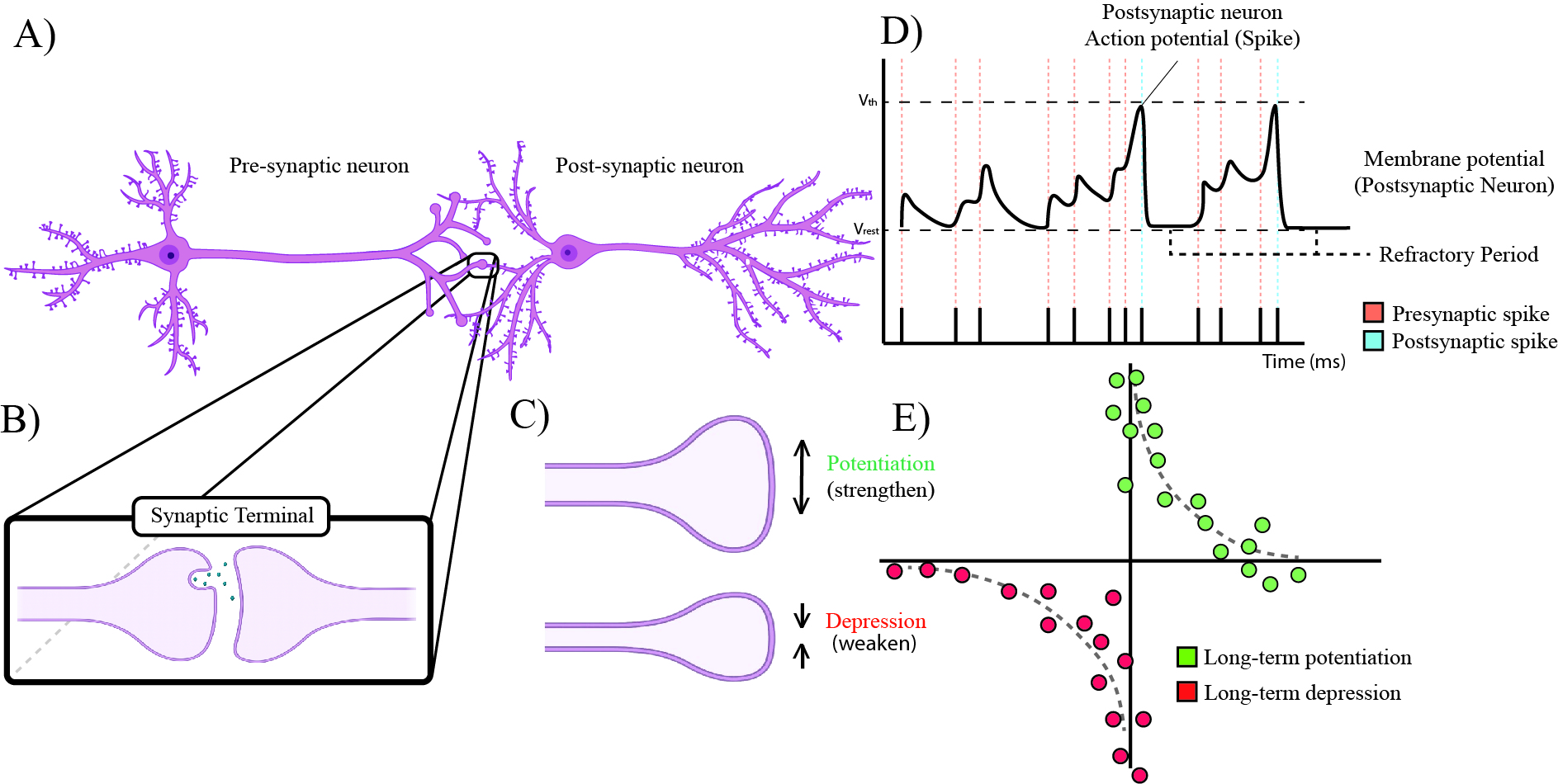}
\caption{Graphical depiction of long-term potentiation (LTP) and depression (LTD) at the synapse biological neurons. \textit{\textbf{A.}} Synaptically connected pre- and post-synaptic neurons. \textit{\textbf{B.}} Synaptic terminal, the connection point between neurons. \textit{\textbf{C.}} Synaptic growth (LTP) and synaptic weakening (LTD). \textit{\textbf{D.}} \textit{Top.} Membrane potential dynamics in the axon hillock of the neuron. \textit{Bottom.} Pre- and post-synaptic spikes. \textbf{\textit{E.}} Spike-timing dependent plasticity curve depicting experimental recordings of LTP and LTD.}
\end{figure*}

%
%Learning is a characteristic ability of biological systems that allows them to adapt to changing environments. These characteristics are primarily enabled by the brain, an intricate system which is constantly undergoing change in order to shape animal behavior. For over half of a century scientists have been trying to develop artificial systems that demonstrate the remarkable abilities of the brain. These attempts did not demonstrate success until the most recent generation of artificial intelligence (AI) systems. These systems have solved many challenging problems such as generating images from speech, controlling complex robotic systems, and mastering games like chess and Go.

%The current paradigm of AI places significant emphasis on results instead of on the learning process. and overlooks a critical characteristic of human learning: that it is robust to changing tasks and sequential experience.

While ANNs have made significant advancements in various fields, there are still major limitations in their ability to continuously learn and adapt like biological brains\cite{kirkpatrick2017overcoming, parisi2019continual, kudithipudi2022biological}. Unlike current models of machine intelligence, animals can learn throughout their entire lifespan, which is essential for stable adaptation to changing environments. This ability, known as lifelong learning, remains a significant challenge for artificial intelligence, which primarily optimizes problems consisting of fixed labeled datasets, causing it to struggle generalizing to new tasks or retain information across repeated learning iterations\cite{parisi2019continual}. Addressing this challenge is an active area of research, and the potential implications of developing AI with lifelong learning abilities could have far-reaching impacts across multiple domains.

In this paper, we offer a unique review that seeks to identify the mechanisms of the brain that have inspired current artificial intelligence algorithms. To better understand the biological processes underlying natural intelligence, the first section will explore the low-level components that shape neuromodulation, from synaptic plasticity, to the role of local and global dynamics that shape neural activity. This will be related back to ANNs in the third section, where we compare and contrast ANNs with biological neural systems. This will give us a logical basis that seeks to justify why the brain has more to offer AI, beyond the inheritance of current artificial models. Following that, we will delve into algorithms of artificial learning that emulate these processes to improve the capabilities of AI systems. Finally, we will discuss various applications of these AI techniques in real-world scenarios, highlighting their potential impact on fields such as robotics, lifelong learning, and neuromorphic computing. By doing so, we aim to provide a comprehensive understanding of the interplay between learning mechanisms in the biological brain and artificial intelligence, highlighting the potential benefits that can arise from this synergistic relationship. We hope our findings will encourage a new generation of brain-inspired learning algorithms.

\section*{Processes that support learning in the brain}

A grand effort in neuroscience aims at identifying the underlying processes of learning in the brain. Several mechanisms have been proposed to explain the biological basis of learning at varying levels of granularity--from the synapse to population-level activity. However, the vast majority of biologically plausible models of learning are characterized by \textit{plasticity} that emerges from the interaction between local and global events\cite{ho2011cell}.  Below, we introduce various forms of plasticity and how these processes interact in more detail.

\paragraph{Synaptic plasticity} Plasticity in the brain refers to the capacity of experience to modify the function of neural circuits. The plasticity of synapses specifically refers to the modification of the strength of synaptic transmission based on activity and is currently the most widely investigated mechanism by which the brain adapts to new information\cite{citri2008synaptic, abraham2019plasticity}. There are two broader classes of synaptic plasticity: short- and long-term plasticity. Short-term plasticity acts on the scale of tens of milliseconds to minutes and has an important role in short-term adaptation to sensory stimuli and short-lasting memory formation\cite{zucker2002short}. Long-term plasticity acts on the scale of minutes to more, and is thought to be one of the primary processes underlying long-term behavioral changes and memory storage\cite{yuste2001morphological}.

%The dominating theory of learning is that synapses are the primary site of information storage in the brain.

\paragraph{Neuromodulation} In addition to the plasticity of synapses, another important mechanism by which the brain adapts to new information is through neuromodulation \cite{fremaux2016neuromodulated, gerstner2018eligibility, brzosko2019neuromodulation}. Neuromodulation refers to the regulation of neural activity by chemical signaling molecules, often referred to as neurotransmitters or hormones. These signaling molecules can alter the excitability of neural circuits and the strength of synapses, and can have both short- and long-term effects on neural function. Different types of neuromodulation have been identified, including acetylcholine, dopamine, and serotonin, which have been linked to various functions such as attention, learning, and emotion \cite{mccormick2020neuromodulation}. Neuromodulation has been suggested to play a role in various forms of plasticity, including short-\cite{zucker2002short} and long-term plasticity\cite{brzosko2019neuromodulation}.

\paragraph{Metaplasticity} The ability of neurons to modify both their function and structure based on activity is what characterizes synaptic plasticity. These modifications which occur at the synapse must be precisely organized so that changes occurs at the right time and by the right quantity. This regulation of plasticity is referred to as \textit{metaplasticity}, or the 'plasticity of synaptic plasticity,' and plays a vital role in safeguarding the constantly changing brain from its own saturation\cite{abraham1996metaplasticity, abraham2008metaplasticity, yger2015models}. Essentially, metaplasticity alters the ability of synapses to generate plasticity by inducing a change in the physiological state of neurons or synapses. Metaplasticity has been proposed as a fundamental mechanism in memory stability, learning, and regulating neural excitability. While similar, metaplasticity can be distinguished from neuromodulation, with metaplastic and neuromodulatory events often overlapping in time during the modification of a synapse.

\paragraph{Neurogenesis} The process by which newly formed neurons are integrated into existing neural circuits is referred to as \textit{neurogenesis}. Neurogenesis is most active during embryonic development, but is also known to occur throughout the adult lifetime, particularly in the subventricular zone of the lateral ventricles \cite{lim2016adult}, the amygdala \cite{roeder2022evidence}, and in the dentate gyrus of the hippocampal formation \cite{kuhn1996neurogenesis}. In adult mice, neurogenesis has been demonstrated to increase when living in enriched environments versus in standard laboratory conditions \cite{kempermann1998experience}. Additionally, many environmental factors such as exercise\cite{van2005exercise, nokia2016physical} and stress\cite{kirby2013acute, baik2020intermittent} have been demonstrated to change the rate of neurogenesis in the rodent hippocampus. Overall, while the role of neurogenesis in learning is not fully understood, it is believe to play an important role in supporting learning in the brain.

\paragraph{Glial Cells} Glial cells, or neuroglia, play a vital role in supporting learning and memory by modulating neurotransmitter signaling at synapses, the small gaps between neurons where neurotransmitters are released and received\cite{todd2006glial}. Astrocytes, one type of glial cell, can release and reuptake neurotransmitters, as well as metabolize and detoxify them. This helps to regulate the balance and availability of neurotransmitters in the brain, which is essential for normal brain function and learning\cite{chung2015astrocytes}. Microglia, another type of glial cell, can also modulate neurotransmitter signaling and participate in the repair and regeneration of damaged tissue, which is important for learning and memory\cite{cornell2022microglia}. In addition to repair and modulation, structural changes in synaptic strength require the involvement of different types of glial cells, with the most notable influence coming from astrocytes\cite{chung2015astrocytes}. However, despite their crucial involvement, we have yet to fully understand the role of glial cells. Understanding the mechanisms by which glial cells support learning at synapses are important areas of ongoing research. 

%https://www.nature.com/articles/s41593-018-0286-y

%https://www.nature.com/articles/s41598-022-10466-8

%\subsection*{Additional forms of neuroplasticity} The various forms of plasticity that occur in the brain (e.g. synaptic plasticity and neurogenesis) are often broadly encapsulated under the term \textit{neuroplasticity}. This term is a generally encompassing term referring to the brain's ability to modify structure and function from experience. While it also includes previously discussed forms of plasticity, there are also many other forms of experience-dependent change that occur at the site of neurons and glial cells or in broader neural systems. 
%https://www.frontiersin.org/articles/10.3389/fpsyg.2017.01657/full#:~:text=Neuroplasticity%20can%20be%20viewed%20as,and%20in%20response%20to%20experience.

%\noindent
%\textbf{Architectural priors.} Learning at the evolutionary scale.

\section*{Deep neural networks and plasticity}

\subsection*{Artificial and spiking neural networks}
Artificial neural networks have played a vital role in machine learning over the past several decades. These networks have catalyzed tremendous progress toward solving a variety of challenging problems. Many of the most impressive accomplishments in AI have been realized through the use of large ANNs trained on tremendous amounts of data. While there have been many technical advancements, many of the accomplishments in AI can be explained by innovations in computing technology, such as large-scale GPU accelerators and the accessibility of data. While the application of large-scale ANNs have led to major innovations, there do exist many challenges ahead. A few of the most pressing practical limitations of ANNs is that they are not efficient in terms of power consumption and they are not very good at processing dynamic and noisy data. In addition, ANNs are not able to learn beyond their training period (e.g. during deployment) from which data assumes an independent and identically distributed (IID) form without time, which does not reflect physical reality where information is highly temporally and spatially correlated. These limitations have led to their application requiring vast amounts of energy when deployed in large-scale settings\cite{desislavov2021compute} and has also presented challenges toward integration into edge computing devices, such as robotics and wearable devices\cite{daghero2021energy}.

Looking toward neuroscience for a solution, researchers have been exploring spiking neural networks (SNNs) as an alternative to ANNs\cite{pfeiffer2018deep}. SNNs are a class of ANNs that are designed to more closely resemble the behavior of biological neurons. The primary difference between ANNs and SNNs is the idea that SNNs incorporate the notion of timing into their communication. Spiking neurons accumulate information across time from connected (presynaptic) neurons (or via sensory input) in the form of a membrane potential. Once a neuron's membrane potential surpasses a threshold value, it fires a binary "spike" to all of its outgoing (postsynaptic) connections. Spikes have been theoretically demonstrated to contain more information than rate-based representations of information (such as in ANNS) despite being both binary and sparse in time\cite{maass1997networks}. Additionally, modelling studies have shown advantages of SNNs, such as better energy efficiency, the ability to process noisy and dynamic data, and the potential for more robust and fault-tolerant computing\cite{schuman2022opportunities}. These benefits are not solely attributed to their increased biological plausibility, but also to the unique properties of spiking neural networks that distinguish them from conventional artificial neural networks. A simple working model of a leaky integrate-and-fire neuron is described below:

\begin{equation*}
\tau_m \frac{dV}{dt} = E_L - V(t) + R_m I_{inj}(t)
\end{equation*}

where $V(t)$ is the membrane potential at time $t$, $\tau_m$ is the membrane time constant, $E_L$ is the resting potential, $R_m$ is the membrane resistance, $I_{inj}(t)$ is the injected current, $V_{th}$ is the threshold potential, and $V_{reset}$ is the reset potential. When the membrane potential reaches the threshold potential, the neuron spikes and the membrane potential is reset to the reset potential (if $V(t) \geq V_{\text{th}} \text{ then } V(t) \leftarrow V_{\text{reset}}$).

Despite these potential advantages, SNNs are still in the early stages of development, and there are several challenges that need to be addressed before they can be used more widely. One of the most pressing challenges is regarding how to optimize the synaptic weights of these models, as traditional backpropagation-based methods from ANNs fail due to the discrete and sparse nonlinearity. Irrespective of these challenges, there do exist some works that push the boundaries of what was thought possible with modern spiking networks, such as large spike-based transformer models\cite{zhu2023spikegpt}. Spiking models are of great importance for this review since they form the basis of many brain-inspired learning algorithms. 

%One challenge is developing more efficient learning algorithms for SNNs, since the spiking behavior of neurons makes learning more difficult than in ANNs. Another challenge is developing more efficient hardware architectures for SNNs, as current computing hardware is not well-suited for simulating the behavior of large-scale SNNs.

\subsection*{Hebbian and spike-timing dependent plasticity}

Hebbian and spike-timing dependent plasticity (STDP) are two prominent models of synaptic plasticity that play important roles in shaping neural circuitry and behavior. The Hebbian learning rule, first proposed by Donald Hebb in 1949\cite{hebb2005organization}, posits that synapses between neurons are strengthened when they are coactive, such that the activation of one neuron causally leads to the activation of another. STDP, on the other hand, is a more recently proposed model of synaptic plasticity that takes into account the precise timing of pre- and post-synaptic spikes\cite{markram2011history} to determine synaptic strengthening or weakening. It is widely believed that STDP plays a key role in the formation and refinement of neural circuits during development and in the ongoing adaptation of circuits in response to experience. In the following subsection, we will provide an overview of the basic principles of Hebbian learning and STDP.

\paragraph{Hebbian learning} Hebbian learning is based on the idea that the synaptic strength between two neurons should be increased if they are both active at the same time, and decreased if they are not. Hebb suggested that this increase should occur when one cell ``repeatedly  or persistently takes part in firing'' another cell (with causal implications). However this principle is often expressed correlatively, as in the famous aphorism "cells that fire together, wire together" (variously attributed to Siegrid L\"owel\cite{lowel1992selection} or Carla Shatz\cite{shatz1992developing})\footnote{As Hebb himself noted, the general idea has a long history. In their review, Brown and colleagues cite William James: "When two elementary brain-processes have been active together or in immediate succession, one of them, on reoccurring, tends to propagate its excitement into the other."}

Hebbian learning is often used as an unsupervised learning algorithm, where the goal is to identify patterns in the input data without explicit feedback\cite{gerstner2014neuronal}. An example of this process is the Hopfield network, in which large binary patterns are easily stored in a fully-connected recurrent network by applying a Hebbian rule to the (symmetric) weights\cite{hopfield1982neural}. It can also be adapted for use in supervised learning algorithms, where the rule is modified to take into account the desired output of the network. In this case, the Hebbian learning rule is combined with a teaching signal that indicates the correct output for a given input. 

A simple Hebbian learning rule can be described mathematically using the equation:

\begin{equation*}
     \Delta w_{ij} = \eta x_i x_j
\end{equation*}

where $\Delta w_{ij}$ is the change in the weight between neuron $i$ and neuron $j$, $\eta$ is the learning rate, and $x_i$ "activity" in neurons $i$, often thought of as the neuron firing rate. This rule states that if the two neurons are activated at the same time, their connection should be strengthened.

One potential drawback of the basic Hebbian rule is its instability. For example, if $x_i$ and $x_j$ are initially weakly positively correlated, this rule  will increase the weight between the two, which will in turn reinforce the correlation, leading to even larger weight increases, etc. Thus, some form of stabilization is needed.  This can be done simply by bounding the weights, or by  more complex rules that  take into account additional factors such as the history of the pre- and post-synaptic activity or the influence of other neurons in the network (see ref. \cite{vasilkoski2011review} for a practical review of many such rules).

\begin{figure*}
    \centering
\includegraphics[width=0.98\linewidth]{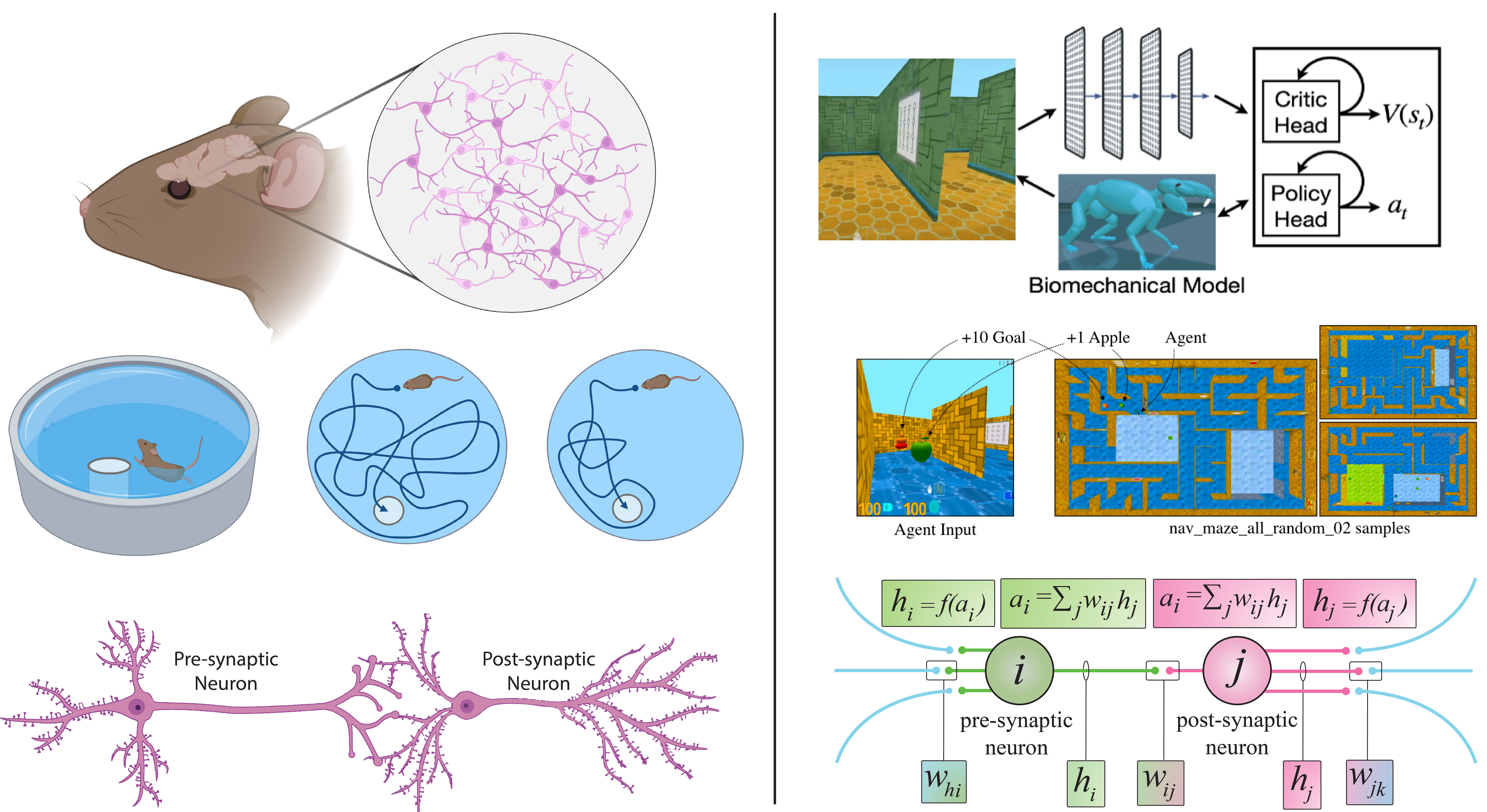}
\caption{\textbf{There are strong parallels between artificial and brain-like learning algorithms.} \textit{\textbf{Left.}} \textit{Top.} Graphical depiction of a rodent and a cluster of interconnected neurons. \textit{Middle.} Rodent is participating in the \textit{Morris water maze} task to test its learning capabilities. \textit{Bottom} A graphical depiction of biological pre- and post-synaptic pyramidal neuron. \textit{\textbf{Right.}} \textit{Top.} A rodent musculoskeletal physics model with artificial neural network policy and critic heads regulating learning and control (see \textit{ref.}\cite{nayebi2022mouse}). \textit{Middle.} A virtual maze environment used for benchmarking learning algorithms (see \textit{ref.}\cite{jaderbergreinforcement}). \textit{Bottom.} An artificial pre- and post-synaptic neuron with forward propagation equations.}
\end{figure*}

\paragraph{Three-factor rules: Hebbian reinforcement learning}

By  incorporating information about rewards, Hebbian learning can also be used for reinforcement learning. An apparently plausible idea is simply to multiply the Hebbian update by the reward directly, as  follows:

\begin{equation*}
     \Delta w_{ij} = \eta x_i x_j R
\end{equation*}

with  R being the reward (for this  time step or for the whole episode). Unfortunately this idea does not produce reliable reinforcement learning. This can be perceived intuitively by noticing that, if $w_{ij}$ is already  at its optimal value, the rule above will still produce a net change and thus drive $w_{ij}$ away from the optimum. 

More formally, as pointed out by Fremaux et al.\cite{fremaux2010functional}, to properly track the actual covariance between inputs,  outputs and rewards, at least one of the terms in the $x_i x_j R$ product must  be centered, that is, replaced by zero-mean fluctuations around its expected value. One possible solution is to center the rewards, by subtracting a baseline from $R$, generally equal  to the expected value of $R$ for this trial. While helpful, in practice this solution is generally insufficient.

A more effective solution is to remove the mean value from the \emph{outputs}. This can be done easily by subjecting neural activations $x_j$ to occasional random perturbations $\Delta x_j$, taken from a suitable zero-centered distribution - and then using the perturbation $\Delta x_j$, rather than the raw post-synaptic activation $x_j$, int he three-factor product:

\begin{equation*}
    \Delta w_{ij} = \eta x_i \Delta x_j R
\end{equation*}
 
This is the so-called "node perturbation" rule proposed by Fiete and Seung\cite{fiete2006gradient,fiete2007model}. Intuitively, notice that the effect of the $x_i \Delta x_j$ increment  is to push future $x_j$ responses (when encountering the same $x_i$ input) in the direction of the perturbation: larger if the perturbation was positive, smaller if the perturbation was negative. Multiplying this shift by $R$ results in pushing future responses towards the perturbation if $R$ was positive, and away from it if $R$ was negative. Even if $R$ is not zero-mean, the net effect (in expectation) will still be to drive $w_{ij}$ towards higher $R$, though the variance will be higher.

This rule turns out to implement the REINFORCE algorithm (Williams' original paper\cite{williams1992simple} actually proposes an algorithm which is exactly node-perturbation for spiking stochastic neurons), and thus estimates the theoretical gradient of $R$ over $w_{ij}$. It an also be implemented in a biologically plausible manner, allowing recurrent networks to learn non-trivial cognitive or motor  tasks from sparse, delayed rewards\cite{miconi2017biologically}.

\paragraph{Spike-timing dependent plasticity}  Spike-timing dependent plasticity (STDP) is a theoretical model of synaptic plasticity that allows the strength of connections between neurons to be modified based on the relative timing of their spikes. Unlike the Hebbian learning rule, which relies on the simultaneous activation of pre- and post-synaptic neurons, STDP takes into account the precise timing of the pre- and post-synaptic spikes. Specifically, STDP suggests that if a presynaptic neuron fires just before a postsynaptic neuron, the connection between them should be strengthened. Conversely, if the postsynaptic neuron fires just before the presynaptic neuron, the connection should be weakened.

STDP has been observed in a variety of biological systems, including the neocortex, hippocampus, and cerebellum. The rule has been shown to play a crucial role in the development and plasticity of neural circuits, including learning and memory processes. STDP has also been used as a basis for the development of artificial neural networks, which are designed to mimic the structure and function of the brain.

The mathematical equation for STDP is more complex than the Hebbian learning rule and can vary depending on the specific implementation. However, a common formulation is:

\begin{equation*}
\Delta w_{ij} = \begin{cases}
A_+ \exp(-\Delta t/\tau_+) & \text{if } \Delta t > 0 \\
-A_{-} \exp(\Delta t / \tau_{-}) & \text{if } \Delta t < 0
\end{cases}
\end{equation*}

where $\Delta w_{ij}$ is the change in the weight between neuron $i$ and neuron $j$, $\Delta t$ is the time difference between the pre- and post-synaptic spikes, $A_+$ and $A_-$ are the amplitudes of the potentiation and depression, respectively, and $\tau_+$ and $\tau_-$ are the time constants for the potentiation and depression, respectively. This rule states that the strength of the connection between the two neurons will be increased or decreased depending on the timing of their spikes relative to each other.

\section*{Processes that support learning in artificial neural networks}

There are two primary approaches for weight optimization in artificial neural networks: error-driven global learning and brain-inspired local learning. In the first approach, the network weights are modified by driving a global error to its minimum value. This is achieved by delegating error to each weight and synchronizing modifications between each weight. In contrast, brain-inspired local learning algorithms aim to learn in a more biologically plausible manner, by modifying weights from dynamical equations using locally available information. Both optimization approaches have unique benefits and drawbacks. In the following sections we will discuss the most utilized form of error-driven global learning, backpropagation, followed by in-depth discussions of brain-inspired local algorithms. It is worth mentioning that these two approaches are not mutually exclusive, and will often be integrated in order to compliment their respective strengths\cite{whittington2020tolman, bellec2020solution, schmidgall2021spikepropamine, schmidgall17meta}.

\subsection*{Backpropagation} Backpropagation is a powerful error-driven global learning method which changes the weight of connections between neurons in a neural network to produce a desired target behavior\cite{rumelhart1986learning}. This is accomplished through the use of a quantitative metric (an objective function) that describes the quality of a behavior given sensory information (e.g. visual input, written text, robotic joint positions). The backpropagation algorithm consists of two phases: the forward pass and the backward pass. In the forward pass, the input is propagated through the network, and the output is calculated. During the backward pass, the error between the predicted output and the "true" output is calculated, and the gradients of the loss function with respect to the weights of the network are calculated by propagating the error backwards through the network. These gradients are then used to update the weights of the network using an optimization algorithm such as stochastic gradient descent. This process is repeated for many iterations until the weights converge to a set of values that minimize the loss function.

Lets take a look at a brief mathematical explanation of backpropagation. First, we define a desired loss function, which is a function of the network's outputs and the true values:

\begin{equation*}
    L(y, \hat{y}) = \frac{1}{2} \sum_i (y_i - \hat{y}_i)^2
\end{equation*}

where $y$ is the true output and $\hat{y}$ is the network's output. In this case we are minimizing the squared error, but could very well optimize for any smooth and differentiable loss function.

Next, we use the chain rule to calculate the gradient of the loss with respect to the weights of the network. Let $w_{ij}^l$ be the weight between neuron $i$ in layer $l$ and neuron $j$ in layer $l+1$, and let $a_i^l$ be the activation of neuron $i$ in layer $l$. Then, the gradients of the loss with respect to the weights are given by:

\begin{equation*}
    \frac{\partial L}{\partial w_{ij}^l} = \frac{\partial L}{\partial a_j^{l+1}} \frac{\partial a_j^{l+1}}{\partial z_j^{l+1}} \frac{\partial z_j^{l+1}}{\partial w_{ij}^l}
\end{equation*}

where $z_j^{l+1}$ is the weighted sum of the inputs to neuron $j$ in layer $l+1$. We can then use these gradients to update the weights of the network using gradient descent:

\begin{equation*}
w_{ij}^l = w_{ij}^l - \alpha \frac{\partial L}{\partial w_{ij}^l}
\end{equation*}

where $\alpha$ is the learning rate. By repeatedly calculating the gradients and updating the weights, the network gradually learns to minimize the loss function and make more accurate predictions. In practice, gradient descent methods are often combined with approaches to incorporate momentum in the gradient estimate, which has been shown to significantly improve generalization\cite{ruder2016overview}.

The impressive accomplishments of backpropagation have led neuroscientists to investigate whether it can provide a better understanding of learning in the brain. While it remains debated as to whether backpropagation variants could occur in the brain \cite{lillicrap2020backpropagation, whittington2019theories}, it is clear that backpropagation in its current formulation is biologically implausible. Alternative theories suggest complex feedback circuits or the interaction of local activity and top-down signals (a "third-factor") could support a similar form of backprop-like learning\cite{lillicrap2020backpropagation}.

Despite its impressive performance there are still fundamental algorithmic challenges that follow from repeatedly applying backpropagation to network weights. One such challenge is a phenomenon known as catastrophic forgetting, where a neural network forgets previously learned information when training on new data\cite{kirkpatrick2017overcoming}. This can occur when the network is fine-tuned on new data or when the network is trained on a sequence of tasks without retaining the knowledge learned from previous tasks. Catastrophic forgetting is a significant hurdle for developing neural networks that can continuously learn from diverse and changing environments. Another challenge is that backpropagation requires backpropagating information through all the layers of the network, which can be computationally expensive and time-consuming, especially for very deep networks. This can limit the scalability of deep learning algorithms and make it difficult to train large models on limited computing resources. Nonetheless, backpropagation has remained the most widely used and successful algorithm for applications involving artificial neural networks

%The optimization framework of artificial neural networks has been proposed as a paradigm to provide new insights into how the brain learns\cite{hennig2021learning}.

\subsection*{Evolutionary and genetic algorithms} 

Another class of global learning algorithms that has gained significant attention in recent years are evolutionary and genetic algorithms. These algorithms are inspired by the process of natural selection and, in the context of ANNs, aim to optimize the weights of a neural network by mimicking the evolutionary process. In \textit{genetic algorithms}\cite{holland1992genetic}, a population of neural networks is initialized with random weights, and each network is evaluated on a specific task or problem. The networks that perform better on the task are then selected for reproduction, whereby they produce offspring with slight variations in their weights. This process is repeated over several generations, with the best-performing networks being used for reproduction, making their behavior more likely across generations. \textit{Evolutionary algorithms} operate similarly to genetic algorithms but use a different approach by approximating a stochastic gradient\cite{de2016evolutionary, salimans2017evolution}. This is accomplished by perturbing the weights and combining the network objective function performances to update the parameters. This results in a more global search of the weight space that can be more efficient at finding optimal solutions compared to local search methods like backpropagation\cite{zhang2017relationship}.

One advantage of these algorithms is their ability to search a vast parameter space efficiently, making them suitable for problems with large numbers of parameters or complex search spaces. Additionally, they do not require a differentiable objective function, which can be useful in scenarios where the objective function is difficult to define or calculate (e.g. spiking neural networks). However, these algorithms also have some drawbacks. One major limitation is the high computational cost required to evaluate and evolve a large population of networks. Another challenge is the potential for the algorithm to become stuck in local optima or to converge too quickly, resulting in suboptimal solutions. Additionally, the use of random mutations can lead to instability and unpredictability in the learning process.

Regardless, evolutionary and genetic algorithms have shown promising results in various applications, particularly when optimizing non-differentiable and non-trivial parameter spaces. Ongoing research is focused on improving the efficiency and scalability of these algorithms, as well as discovering where and when it makes sense to use these approaches instead of gradient descent.

\section*{Brain-inspired representations of learning in artificial neural networks}

\subsection*{Local learning algorithms}

Unlike global learning algorithms such as backpropagation, which require information to be propagated through the entire network, local learning algorithms focus on updating synaptic weights based on local information from nearby or synaptically connected neurons. These approaches are often strongly inspired by the plasticity of biological synapses. As will be seen, by leveraging local learning algorithms, ANNs can learn more efficiently and adapt to changing input distributions, making them better suited for real-world applications. In this section, we will review recent advances in brain-inspired local learning algorithms and their potential for improving the performance and robustness of ANNs.

\subsection*{Backpropagation-derived local learning}

Backpropagation-derived local learning algorithms are a class of local learning algorithms that attempt to emulate the mathematical properties of backpropagation. Unlike the traditional backpropagation algorithm, which involves propagating error signals back through the entire network, backpropagation-derived local learning algorithms update synaptic weights based on local error gradients computed using backpropagation. This approach is computationally efficient and allows for online learning, making it suitable for applications where training data is continually arriving. 

One prominent example of backpropagation-derived local learning algorithms is the Feedback Alignment (FA) algorithm \cite{lillicrap2014random, nokland2016direct}, which replaces the weight transport matrix used in backpropagation with a fixed random matrix, allowing the error signal to propagate from direct connections thus avoiding the need for backpropagating error signals. A brief mathematical description of feedback alignment is as follows: let $\textit{w}^{out}$ be the weight matrix connecting the last layer of the network to the output, and $\textit{w}^{in}$ be the weight matrix connecting the input to the first layer. In Feedback Alignment, the error signal is propagated from the output to the input using the fixed random matrix $\textit{B}$, rather than the transpose of $\textit{w}^{out}$. The weight updates are then computed using the product of the input and the error signal, $\Delta w^{in} = -\eta x z$ where $\textit{x}$ is the input, $\eta$ is the learning rate, and $\textit{z}$ is the error signal propagated backwards through the network, similar to traditional backpropagation.

%Δw=−ηxz

Direct Feedback Alignment\cite{nokland2016direct} (DFA) simplifies the weight transport chain compared with FA by directly connecting the output layer error to each hidden layer. The Sign-Symmetry (SS) algorithm is similar to FA except the feedback weights symmetrically share signs. While FA has exhibited impressive results on small datasets like MNIST and CIFAR, their performance on larger datasets such as ImageNet is often suboptimal\cite{bartunov2018assessing}. On the other hand, recent studies have shown that the SS algorithm algorithm is capable of achieving comparable performance to backpropagation, even on large-scale datasets\cite{xiao2018biologically}.

%Eligibility propagation\cite{bellec2019eligibility, bellec2020solution} (e-prop) extends the idea of feedback alignment for spiking neural networks, combining the advantages of both traditional error backpropagation and biologically plausible learning rules, such as spike-timing-dependent plasticity (STDP). The e-prop algorithm uses a variant of the backpropagation algorithm to calculate the gradient of the network's output with respect to the synaptic weights, but instead of propagating error signals back through the network, it computes eligibility traces at each synapse. The eligibility trace is a measure of the contribution of a synapse to the network's output and is updated based on the temporal coincidence between the pre- and postsynaptic spikes\cite{gerstner2018eligibility}. The eligibility trace is then used to update the synaptic weights in a biologically plausible manner, such as through STDP.

%%% Thomas
Eligibility propagation\cite{bellec2019eligibility, bellec2020solution} (e-prop) extends the idea of feedback alignment for spiking neural networks, combining the advantages of both traditional error backpropagation and biologically plausible learning rules, such as spike-timing-dependent plasticity (STDP). For each synapse,  the e-prop algorithm computes and maintains an eligibility trace $e_{ji}(t) = \frac{dz_j(t)}{dW_{ji}}$.  Eligibility traces measure the total contribution of this synapse to the neuron's current output,  taking into account all past inputs \cite{gerstner2018eligibility}. This can be computed and updated in a purely forward manner, without backward passes. This eligibility trace is then multiplied by  an estimate of the gradient of the error over the neuron's output $L_j(t)= \frac{dE(t)}{dz_j(t)}$. to obtain the actual weight gradient  $\frac{dE(t)}{dW_{ji}}$. $L_j(t)$ itself is  computed from the error at the output neurons,  either by using symmetric feedback weights or by using fixed feedback weights, as in feedback alignment. A possible drawback of e-prop is that it requires a real-time error signal $L_t$  at each point in time, since it only takes into account past events and is blind to future errors. In particular, it cannot  learn from delayed error signals that extend beyond the time scales of individual neurons (including short-term adaptation)\cite{bellec2020solution}, in contrast with methods like REINFORCE and node-perturbation.

In the work of ref. \cite{liu2021cell, liu2022biologically} a normative theory for synaptic learning based on recent genetic findings\cite{smith2019single} of neuronal signaling architectures is demonstrated. They propose that neurons communicate their contribution to the learning outcome to nearby neurons via cell-type-specific local neuromodulation, and that neuron-type diversity and neuron-type-specific local neuromodulation may be critical pieces of the biological credit-assignment puzzle. In this work, the authors instantiate a simplified computational model based on eligibility propagation to explore this theory and show that their model, which includes both dopamine-like temporal difference and neuropeptide-like local modulatory signaling, leads to improvements over previous methods such as e-prop and feedback alignment.

\paragraph{Generalization properties} Techniques in deep learning have made tremendous strides toward understanding the generalization of their learning algorithms. A particularly useful discovery was that flat minima tend to lead to better generalization \cite{hochreiter1997flat}. What is meant by this is that, given a perturbation $\epsilon$ in the parameter space (synaptic weight values) more significant performance degradation is observed around \textit{narrower} minima. Learning algorithms that find \textit{flatter} minima in parameter space ultimately lead to better generalization.
%PARDIS: Studies have shown that learning algorithms that find flatter minima in the parameter space tend to lead to better generalization, or the ability of a model to perform well on unseen data. This is because perturbations in the parameter space are more likely to result in similar objective function values, making the model more robust to small changes in the weights. On the other hand, narrow minima have steep walls around them, and small perturbations can result in a significant increase in the objective function value, leading to overfittin% 

Recent work has explored the generalization properties exhibited by (brain-inspired) backpropagation-derived local learning rules\cite{liu2022beyond}. Compared with backpropagation through time, backpropagation-derived local learning rules exhibit worse and more variable generalization which does not improve by scaling the step size due to the gradient approximation being poorly aligned with the true gradient. While it is perhaps unsurprising that \textit{local approximations} of an optimization process are going to have worse generalization properties than their complete counterpart, this work opens the door toward asking new questions about what the best approach toward designing brain-inspired learning algorithms is. It also opens the question as to whether backpropagation-derived local learning rules are even worth exploring given that they are fundamentally going to exhibit \textit{sub-par} generalization.

%Nonetheless, despite their promising performance, backpropagation-derived local learning algorithms have limitations, such as the need for carefully tuned hyperparameters and the potential for instability when training deep neural networks. In addition, despite some of these algorithms being local and online, backpropagation-based plasticity methods still struggle with the same issues as backpropagation, like catastrophic forgetting, since they are still fundamentally approximating gradient descent. Nonetheless, these methods present themselves as a promising new direction for brain-inspired learning in artificial neural networks.

% PARDIS: Recent research has compared the generalization properties of brain-inspired local learning rules derived from backpropagation with those of complete backpropagation through time. The findings have shown that local learning rules exhibit worse and more variable generalization, even when the step size is scaled, due to poor alignment between the gradient approximation and the true gradient. Although these algorithms offer a promising direction for brain-inspired learning in artificial neural networks, their limitations, including the need for carefully tuned hyperparameters and the potential for instability when training deep neural networks, are concerning. Furthermore, the poor generalization of these algorithms raises questions about the best approach to designing brain-inspired learning algorithms.

In conclusion, while backpropagation-derived local learning rules present themselves as a promising approach to designing brain-inspired learning algorithms, they come with limitations that must be addressed. The poor generalization of these algorithms highlights the need for further research to improve their performance and to explore alternative brain-inspired learning rules. It also opens the question as to whether backpropagation-derived local learning rules are even worth exploring given that they are fundamentally going to exhibit sub-par generalization.

% Meta-learning occurs when one learning system progressively adjusts the operation of a second learning system, such that the latter operates with increasing speed and efficiency

\subsection*{Meta-optimized plasticity rules}
% Louis main section -- potentially requires further separation in sections or paragraph headings
% Probably paragraph headings like in previous sections

% Structure brainstorming
% * Learned plasticity / fast weights from cs and neuros
% * historic background (in cs: schmiduber, bengio, stanley)
% * example: differentiable plasticity schmidgal / miconi
% * memory-based meta-learning, in-context learning (hochreiter, wang, duan, brown)
% * How are these connected? parameter sharing (& sparsity + multiplicative interactions)
% * generalization + general-purposeness: VSML + SymLA
% * connection to transformers: outer-products, self-attention (schlag)
% * introducing self-ref works (schmidhuber + irie + kirsch)
% * evolutionary bias or SGD architectures (metz + lange)
% * self-ref fast weights meta-learn initialization (irie)
% * eliminating meta optimization (kirsch)

% TODO: Maybe discuss how to meta-optimize: Backprop vs Evolution vs RL etc
% TODO: include gradient-based meta learners? Probably too tangential.

% Background meta learning
Meta-optimized plasticity rules offer an effective balance between error-driven global learning and brain-inspired local learning.
Meta-learning can be defined as automation of the search for learning algorithms themselves, where, instead of relying on human engineering to describe a learning algorithm, a search process to find that algorithm is employed\cite{schmidhuber1987evolutionary}.
The idea of meta-learning naturally extends to brain-inspired learning algorithms, such that the brain-inspired mechanism of learning itself can be optimized thereby allowing for \textit{discovery} of more efficient learning without manual tuning of the rule. In the following section, we discuss various aspects of this research starting with differentiably optimized synaptic plasticity rules.

% including differentiability optimized synaptic plasticity rules in ANNs and SNNs, plasticity in recurrent neural networks and transformers,
%When the meta-learner is set up to directly output the synaptic weight updates, this is known as learned plasticity\cite{bengio1990learning} or fast weight programmers\cite{schmidhuber1992learning,schmidhuber1993reducing}.
% In the following, we discuss various aspects of this research, including differentiability, 
% * Learned plasticity / fast weights from cs and neuros
% * historic background (in cs: schmiduber, bengio, stanley)

% * for a given task sounds like it doesn't generalize very well
% * there are some discovered rules which are actually quite generalizing
% Meta-optimized plasticity rules take advantage of this concept by incorporating a meta-learning process that learns the optimal plasticity rules for a given task.
% This approach can efficiently search through a large space of potential plasticity rules and parameters to find the most effective ones for a given task.

\begin{figure*}
    \centering
\includegraphics[width=0.95\linewidth]{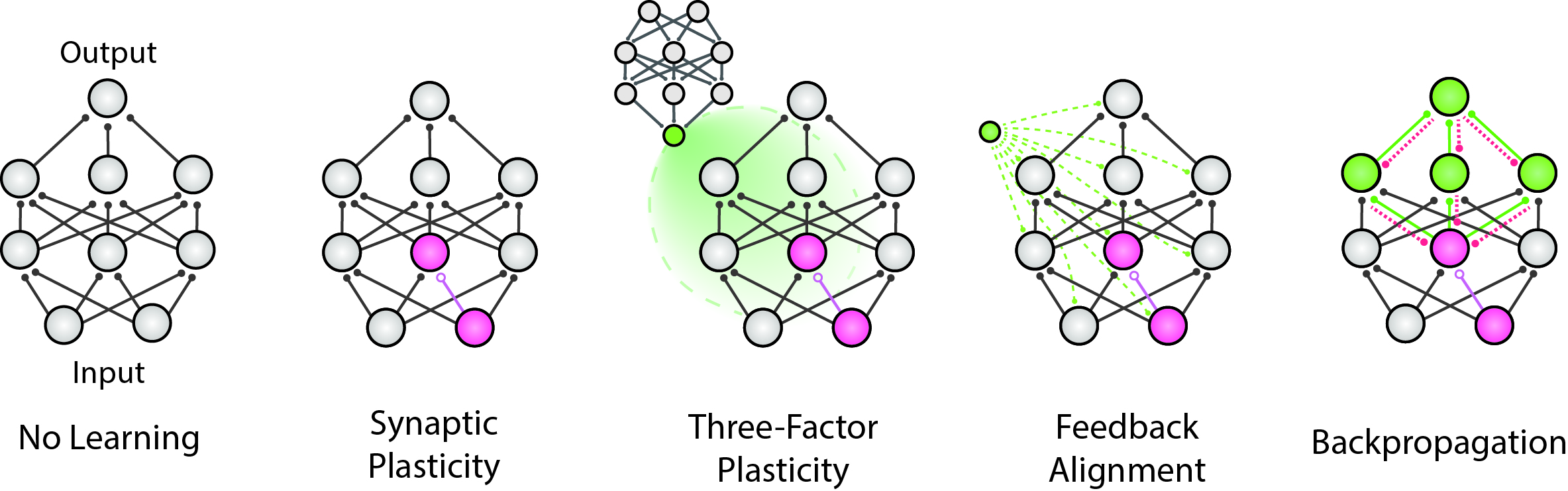}
\caption{A feedforward neural network computes an output given an input by propagating the input information downstream. The precise value of the output is determined by the weight of synaptic coefficients. To improve the output for a task given an input, the synaptic weights are modified. \textit{\textbf{Synaptic Plasticity}} algorithms represent computational models that emulate the brain's ability to strengthen or weaken synapses-connections between neurons-based on their activity, thereby facilitating learning and memory formation. \textit{\textbf{Three-Factor Plasticity}} refers to a model of synaptic plasticity in which changes to the strength of neural connections are determined by three factors: pre-synaptic activity, post-synaptic activity, and a modulatory signal, facilitating more nuanced and adaptive learning processes. The \textit{\textbf{Feedback Alignment}} algorithm is a learning technique in which artificial neural networks are trained using random, fixed feedback connections rather than symmetric weight matrices, demonstrating that successful learning can occur without precise backpropagation. \textit{\textbf{Backpropagation}} is a fundamental algorithm in machine learning and artificial intelligence, used to train neural networks by calculating the gradient of the loss function with respect to the weights in the network.}
\end{figure*}

% Differentiable plasticity
% * example: differentiable plasticity schmidgal / miconi
\paragraph{Differentiable plasticity}
One instantiation of this principle in the literature is \textit{differentiable plasticity}, which is a framework that focuses on optimizing synaptic plasticity rules in neural networks through gradient descent\cite{miconi2018differentiable, miconi2020backpropamine}.
In these rules, the plasticity rules are described in such a way that the parameters governing their dynamics are differentiable, allowing for backpropagation to be used for meta-optimization of the plasticity rule parameters (e.g. the $\eta$ term in the simple hebbian rule or the $A_{+}$ term in the STDP rule).
This allows the weight dynamics to precisely solve a task that requires the weights to be optimized during execution time, referred to as intra-lifetime learning.

Differentiable plasticity rules are also capable of the differentiable optimization of neuromodulatory dynamics\cite{miconi2020backpropamine, schmidgall2021spikepropamine}. This framework includes two main variants of neuromodulation: global neuromodulation, where the direction and magnitude of weight changes is controlled by a network-output-dependent global parameter, and retroactive neuromodulation, where the effect of past activity is modulated by a dopamine-like signal within a short time window.
This is enable by the use of eligibility traces, which are used to keep track of which synapses contributed to recent activity, and the dopamine signal modulates the transformation of these traces into actual plastic changes.

Methods involving differentiable plasticity have seen improvements in a wide range of applications from sequential associative tasks\cite{duan2023hebbian}, familiarity detection\cite{tyulmankov2022meta}, and robotic noise adaptation\cite{schmidgall2021spikepropamine}.
This method has also been used to optimize short-term plasticity rules\cite{rodriguez2022short,tyulmankov2022meta} which exhibit improved performance in reinforcement and temporal supervised learning problems. While these methods show much promise, differentiable plasticity approaches take a tremendous amount of memory, as backpropagation is used to optimize multiple parameters \textit{for each synapse} through time. Practical advancements with these methods will likely require parameter sharing\cite{palm2021testing} or a more memory-efficient form of backpropagation\cite{gruslys2016memory}. 

% Spiking neuron versions
\paragraph{Plasticity with spiking neurons}
Recent advances in backpropagating through the non-differentiable part of spiking neurons with surrogate gradients have allowed for differentiable plasticity to be used to optimize plasticity rules in spiking neural networks\cite{schmidgall2021spikepropamine}.
In ref.\cite{schmidgall17meta} the capability of this optimization paradigm is demonstrated through the use of a differentiable spike-timing dependent plasticity rule to enable "learning to learn" on an online one-shot continual learning problem and on an online one-shot image class recognition problem.
A similar method was used to optimize the third-factor signal using the gradient approximation of e-prop as the plasticity rule, introducing a meta-optimization form of e-prop\cite{scherr2020one}.
% ENUs
Recurrent neural networks tuned by evolution can also be used for meta-optimized learning rules.
Evolvable Neural Units\cite{bertens2020network} (ENUs) introduce a gating structure that controls how the input is processed, stored, and dynamic parameters are updated.
This work demonstrates the evolution of individual somatic and synaptic compartment models of neurons and show that a network of ENUs can learn to solve a T-maze environment task, independently discovering spiking dynamics and reinforcement-type learning rules.
% Another method introduces an evolvable neural unit\cite{bertens2020network} (ENU) which optimizes synaptic and neuronal dynamics through the use of a \textit{recurrent neural network} that is tuned via evolutionary algorithms.
% The ENU has a gating structure that enables fine control over how input influences the internal memory state, which in turn controls how that input is processed and how dynamic parameters are updated.
% This architecture allows for the evolution of spike-based behaviour through storing and summing received input into the memory state and then, once some threshold is reached, producing an output spike.
% Using ENUs, this work demonstrates the evolution of individual somatic and synaptic compartment models of neurons and show that a network of ENUs can learn to solve a T-maze environment task, independently discovering spiking dynamics and reinforcement-type learning rules.

\paragraph{Plasticity in RNNs and Transformers}
Independent of research aiming at learning plasticity using update rules, Transformers have recently been shown to be good intra-lifetime learners\cite{brown2020language,garg2022can,kirsch2022general}.
The process of in-context learning  works not through the update of synaptic weights but purely within the network activations.
Like in Transformers, this process can also happen in recurrent neural networks\cite{hochreiter2001learning}.
While in-context learning appears to be a different mechanism from synaptic plasticity, these processes have been demonstrated to exhibit a strong relationship.
One exciting connection discussed in the literature is the realization that parameter-sharing of the meta-learner often leads to the \textit{interpretation of activations as weights}\cite{kirsch2021meta}. 
This demonstrates that, while these models may have fixed weights, they exhibit some of the same learning capabilities as models with plastic weights.
Another connection is that self-attention in the Transformer involves outer and inner products that can be cast as learned weight updates\cite{schlag2021linear} that can even implement gradient descent\cite{akyurek2022learning,von2022transformers}.

\paragraph{Evolutionary and genetic meta-optimization}
Much like differentiable plasticity, evolutionary and genetic algorithms have been used to optimize the parameters of plasticity rules on a variety of applications\cite{soltoggio2018born}, including: adaptation to limb damage on robotic systems\cite{schmidgall2020adaptive, najarro2020meta}.
Recent work has also enabled the optimization of both plasticity coefficients and plasticity rule \textit{equations} through the use of Cartesian genetic programming\cite{10.7554/eLife.66273}, presenting an automated approach for discovering biologically plausible plasticity rules based on the specific task being solved.
In these methods, the genetic or evolutionary optimization process acts similarly to the differentiable process such that it optimizes the plasticity parameters in an outer-loop process, while the plasticity rule optimizes the reward in an inner-loop process. These methods are appealing since they have a much lower memory footprint compared to differentiable methods since they do not require backpropagating errors through time. However, while memory efficient, they often require a tremendous amount of data to get comparable performance to gradient-based methods\cite{pagliuca2020efficacy}.

%It is demonstrated that gradient-descent plasticity and spike-timing dependent plasticity rules can be discovered by the algorithm depending on the task being solved.

\paragraph{Self-referential meta-learning}
While synaptic plasticity has two-levels of learning, the meta-learner, and the discovered learning rule, self-referential meta-learning\cite{schmidhuber1993self,kirsch2022eliminating} extends this hierarchy.
% Self-referential meta-learning\cite{schmidhuber1993self,kirsch2022eliminating} is a type of machine learning approach related to synaptic plasticity.
In plasticity approaches only a subset of the network parameters are updated (e.g. the synaptic weights), whereas the meta-learned update rule remains fixed after meta-optimization.
Self-referential architectures enable a neural network to modify all of its parameters in recursive fashion.
Thus, the learner can also modify the meta-learner.
This in principles allows arbitrary levels of learning, meta-learning, meta-meta-learning, etc.
Some approaches meta-learn the parameter initialization of such a system\cite{schmidhuber1993self,irie2022modern}.
Finding this initialization still requires a hardwired meta-learner.
In other works the network self-modifies in a way that eliminates even this meta-learner\cite{kirsch2022self,kirsch2022eliminating}.
Sometimes the learning rule to be discovered has structural search space restrictions which simplify self-improvement where a gradient-based optimizer can discover itself\cite{metz2021training} or an evolutionary algorithm can optimize itself\cite{lange2022discovering}.
Despite their differences, both synaptic plasticity, as well as self-referential approaches, aim to achieve self-improvement and adaptation in neural networks.

\paragraph{Generalization of meta-optimized learning rules}
The extent to which discovered learning rules generalize to a wide range of tasks is a significant open question--in particular, when should they replace manually derived general-purpose learning rules such as backpropagation?
A particular observation that poses a challenge to these methods is that when the search space is large and few restrictions are put on the learning mechanism\cite{hochreiter2001learning,wang2016learning,duan2016rl}, generalization is shown to become more difficult.
However, toward amending this, in variable shared meta learning\cite{kirsch2021meta} flexible learning rules were parameterized by parameter-shared recurrent neural networks that locally exchange information to implement learning algorithms that generalize across classification problems not seen during meta-optimization.
Similar results have also been shown for the discovery of reinforcement learning algorithms\cite{kirsch2022introducing}.

\section*{Applications of brain-inspired learning}

\paragraph{Neuromorphic Computing}

Neuromorphic computing represents a paradigm shift in the design of computing systems, with the goal of creating hardware that mimics the structure and functionality of the biological brain \cite{schuman2017survey, yang2020neuromorphic, schuman2022opportunities}. This approach seeks to develop artificial neural networks that not only replicate the brain's learning capabilities but also its energy efficiency and inherent parallelism. Neuromorphic computing systems often incorporate specialized hardware, such as neuromorphic chips or memristive devices, to enable the efficient execution of brain-inspired learning algorithms \cite{yang2020neuromorphic}. These systems have the potential to drastically improve the performance of machine learning applications, particularly in edge computing and real-time processing scenarios.

A key aspect of neuromorphic computing lies in the development of specialized hardware architectures that facilitate the implementation of spiking neural networks, which more closely resemble the information processing mechanisms of biological neurons. Neuromorphic systems operate based on the principle of brain-inspired local learning, which allows them to achieve high energy efficiency, low-latency processing, and robustness against noise, which are critical for real-world applications\cite{khacef2022spike}. The integration of brain-inspired learning techniques with neuromorphic hardware is vital for the successful application of this technology.

In recent years, advances in neuromorphic computing have led to the development of various platforms, such as Intel's Loihi\cite{davies2018loihi}, IBM's TrueNorth\cite{akopyan2015truenorth}, and SpiNNaker\cite{painkras2013spinnaker}, which offer specialized hardware architectures for implementing SNNs and brain-inspired learning algorithms. These platforms provide a foundation for further exploration of neuromorphic computing systems, enabling researchers to design, simulate, and evaluate novel neural network architectures and learning rules. As neuromorphic computing continues to progress, it is expected to play a pivotal role in the future of artificial intelligence, driving innovation and enabling the development of more efficient, versatile, and biologically plausible learning systems.

\paragraph{Robotic learning}

Brain-inspired learning in neural networks has the potential to overcome many of the current challenges present in the field of robotics by enabling robots to learn and adapt to their environment in a more flexible way \cite{floreano2014robotics, bing2018survey}. Traditional robotics systems rely on pre-programmed behaviors, which are limited in their ability to adapt to changing conditions. In contrast, as we have shown in this review, neural networks can be trained to adapt to new situations by adjusting their internal parameters based on the data they receive. 

Because of their natural relationship to robotics, brain-inspired learning algorithms have a long history in robotics\cite{floreano2014robotics}. Toward this, synaptic plasticity rules have been introduced for adapting 
robotic behavior to domain shifts such as motor gains and rough terrain\cite{grinke2015synaptic, kaiser2019embodied, schmidgall2021spikepropamine, schmidgallThreeFactor} as well as for obstacle avoidance\cite{arena2010insect, hu2014digital, wang2014mobile} and articulated (arm) control\cite{neymotin2013reinforcement, dura2015cortical}. Brain-inspired learning rules have also been used to explore how learning occurs in the insect brain using robotic systems as an embodied medium\cite{ilg1995learning, ijspeert2014biorobotics, faghihi2017computational, szczecinski2023perspective}. 

Deep reinforcement learning (DRL) represents a significant success of brain-inspired learning algorithms, combining the strengths of neural networks with the theory of reinforcement learning in the brain to create autonomous agents capable of learning complex behaviors through interaction with their environment \cite{botvinick2020deep,arulkumaran2017brief, mnih2015human}. By utilizing a reward-driven learning process emulating the activity of dopamine neurons\cite{watabe2017neural}, as opposed to the minimization of an e.g classification or regression error, DRL algorithms guide robots toward learning optimal strategies to achieve their goals, even in highly dynamic and uncertain environments \cite{kaelbling1996reinforcement, sutton2018reinforcement}. This powerful approach has been demonstrated in a variety of robotic applications, including dexterous manipulation, robotic locomotion \cite{peng2018deepmimic}, and multi-agent coordination \cite{lowe2017multi}.

%Another promising application of brain-inspired learning in robotics is toward improving the performance of robots by enabling them to learn and adapt to new tasks and environments across the duration of their lifetime. 

\paragraph{Lifelong and online learning}

Lifelong and online learning are essential applications of brain-inspired learning in artificial intelligence, as they enable systems to adapt to changing environments and continuously acquire new skills and knowledge\cite{parisi2019continual}. Traditional machine learning approaches, in contrast, are typically trained on a fixed dataset and lack the ability to adapt to new information or changing environments. The mature brain is an incredible medium for lifelong learning, as it is constantly learning while remaining relatively fixed in size across the span of a lifetime\cite{la2020brain}. As this review has demonstrated, neural networks endowed with brain-inspired learning mechanisms, similar to the brain, can be trained to learn and adapt continuously, improving their performance over time.

The development of brain-inspired learning algorithms that enable artificial systems to exhibit this capability has the potential to significantly enhance their performance and capabilities and has wide-ranging implications for a variety of applications. These applications are particularly useful in situations where data is scarce or expensive to collect, such as in robotics\cite{lesort2020continual} or autonomous systems\cite{shaheen2022continual}, as it allows the system to learn and adapt in real-time rather than requiring large amounts of data to be collected and processed before learning can occur. 

One of the primary objectives in the field of lifelong learning is to alleviate a major issue associated with the continuous application of backpropagation on ANNs, a phenomenon known as catastrophic forgetting\cite{kirkpatrick2017overcoming}. Catastrophic forgetting refers to the tendency of an ANN to abruptly forget previously learned information upon learning new data. This happens because the weights in the network that were initially optimized for earlier tasks are drastically altered to accommodate the new learning, thereby erasing or overwriting the previous information. This is because the backpropagation algorithm does not inherently factor in the need to preserve previously acquired information while facilitating new learning. Solving this problem has remained a significant hurdle in AI for decades. We posit that by employing brain-inspired learning algorithms, which emulate the dynamic learning mechanisms of the brain, we may be able to capitalize on the proficient problem-solving strategies inherent to biological organisms.

\paragraph{Toward understanding the brain}
The worlds of artificial intelligence and neuroscience have been greatly benefiting from each other. Deep neural networks, specially tailored for certain tasks, show striking similarities to the human brain in how they handle spatial \cite{banino2018vector, cueva2018emergence, gao2022computational} and visual \cite{schrimpf2018brain, nayebi2021shallow, jacob2021qualitative} information. This overlap hints at the potential of artificial neural networks (ANNs) as useful models in our efforts to better understand the brain's complex mechanics. A new movement referred to as \textit{the neuroconnectionist research programme} \cite{doerig2022neuroconnectionist} embodies this combined approach, using ANNs as a computational language to form and test ideas about how the brain computes. This perspective brings together different research efforts, offering a common computational framework and tools to test specific theories about the brain.

While this review highlights a range of algorithms that imitate the brain's functions, we still have a substantial amount of work to do to fully grasp how learning actually happens in the brain. The use of backpropagation, and backpropagation-like local learning rules, to train large neural networks may provide a good starting point for modelling brain function. Much productive investigation has occurred to see what processes in the brain may operate similarly to  backpropagation\cite{lillicrap2020backpropagation}, leading to new perspectives and theories in neuroscience. Even though backpropagation in its current form might not occur in the brain, the idea that the brain might develop similar internal representations to ANNs despite such different mechanisms of learning is an exciting open question that may lead to a deeper understanding of the brain and of AI.

Explorations are now extending beyond static network dynamics to the networks which unravel a function of time much like the brain. As we further develop algorithms in continual and lifelong learning, it may become clear that our models need to reflect the learning mechanisms observed in nature more closely. This shift in focus calls for the integration of local learning rules—those that mirror the brain's own methods—into ANNs.

We are convinced that adopting more biologically authentic learning rules within ANNs will not only yield the aforementioned benefits, but it will also serve to point neuroscience researchers in the right direction.. In other words, it's a strategy with a two-fold benefit: not only does it promise to invigorate innovation in engineering, but it also brings us closer to unravelling the intricate processes at play within the brain. With more realistic models, we can probe deeper into the complexities of brain computation from the novel perspective of artificial intelligence.

\section*{Conclusion}

%In this review, we investigated the integration of more biologically plausible learning mechanisms into ANNs. Our discussion begins with a concise summary of the processes that underpin learning in the brain. Following this, we offer an examination of the fundamentals of deep neural networks, spiking neural networks, and two central forms of synaptic plasticity. We then provide an overview of the mechanisms that facilitate learning in ANNs, including but not limited to backpropagation and evolutionary algorithms. The subsequent section focuses on an analysis of the processes that foster learning in ANNs, notably backpropagation-driven local learning and meta-optimized plasticity rules. To conclude, we engage in a discussion regarding the applications of brain-inspired learning rules in diverse areas such as neuromorphic computing, robotic learning, and toward furthering our understanding of the brain itself.

In this review, we investigated the integration of more biologically plausible learning mechanisms into ANNs. This further integration presents itself as an important step for both neuroscience and artificial intelligence. This is particularly relevant amidst the tremendous progress that has been made in artificial intelligence with large language models and embedded systems, which are in critical need for more energy efficient approaches for learning and execution. Additionally, while ANNs are making great strides in these applications, there are still major limitations in their ability to adapt like biological brains, which we see as a primary application of brain-inspired learning mechanisms.

As we strategize for future collaboration between neuroscience and AI toward more detailed brain-inspired learning algorithms, it's important to acknowledge that the past influences of neuroscience on AI have seldom been about a straightforward application of ready-made solutions to machines\cite{hassabis2017neuroscience}. More often, neuroscience has stimulated AI researchers by posing intriguing algorithmic-level questions about aspects of animal learning and intelligence. It has provided preliminary guidance towards vital mechanisms that support learning. Our perspective is that by harnessing the insights drawn from neuroscience, we can significantly accelerate advancements in the learning mechanisms used in ANNs. Likewise, experiments using brain-like learning algorithms in AI can accelerate our understanding of neuroscience.

\section*{Acknowledgements}

We thank the OpenBioML collaborate workspace from which several of the authors of this work were connected.
This material is based upon work supported by the National Science Foundation Graduate Research Fellowship under Grant No. DGE2139757.

%Discoveries in neuroscience have demonstrated that synaptic weight potentiation and depression occur even in the absence of activity in dendritic spines.

%\bibliographystyle{zHenriquesLab-StyleBib}
\bibliography{sample}
\end{document}